\documentclass[conference]{IEEEtran}
\IEEEoverridecommandlockouts
\usepackage{cite}
\usepackage{amsmath,amssymb,amsfonts}
\usepackage{algorithmic}
\usepackage{graphicx}
\usepackage{textcomp}
\usepackage{xcolor}
\usepackage{enumitem}
\usepackage{multirow}
\usepackage{array}

\def\BibTeX{{\rm B\kern-.05em{\sc i\kern-.025em b}\kern-.08em
    T\kern-.1667em\lower.7ex\hbox{E}\kern-.125emX}}
\begin{document}

\title{Improving Interpretability in Alzheimer's Prediction via Joint Learning of ADAS-Cog Scores\\

\thanks{https://github.com/mirahhamid/MTL-Improving-Interpretability-in-Alzheimer-s-Prediction-via-Joint-Learning-of-ADAS-Cog-Scores}
}

\author{
\IEEEauthorblockN{ Nur Amirah Abd Hamid}
\IEEEauthorblockA{\textit{School of Digital Science} \\
\textit{Universiti Brunei Darussalam}\\
Bandar Seri Begawan, Brunei Darussalam \\
mirahhamid5496@gmail.com}
\and
\IEEEauthorblockN{ Mohd Shahrizal Rusli}
\IEEEauthorblockA{\textit{Faculty of Artificial Intelligence} \\
\textit{Universiti Teknologi Malaysia}\\
Kuala Lumpur, Malaysia \\
shahrizal@fke.utm.my}
\and
\IEEEauthorblockN{ Muhammad Thaqif Iman Mohd Taufek}
\IEEEauthorblockA{\textit{Universiti Teknologi Malaysia} \\
Kuala Lumpur, Malaysia \\
muhammadthaqifiman@graduate.utm.my}
\and
\IEEEauthorblockN{ Mohd Ibrahim Shapiai}
\IEEEauthorblockA{\textit{Malaysia-Japan International Institute of Technology} \\
\textit{Universiti Teknologi Malaysia}\\
Kuala Lumpur, Malaysia \\
md\_ibrahim83@utm.my}
\and
\IEEEauthorblockN{Daphne Teck Ching Lai}
\IEEEauthorblockA{\textit{School of Digital Science} \\
\textit{Universiti Brunei Darussalam}\\
Bandar Seri Begawan, Brunei Darussalam \\
daphne.lai@ubd.edu.bn}
}

\maketitle

\begin{abstract}
Accurate prediction of clinical scores is critical for early detection and prognosis of Alzheimer’s disease (AD). While existing approaches primarily focus on forecasting the ADAS-Cog global score, they often overlook the predictive value of its sub-scores (13-items), which capture domain-specific cognitive decline. In this study, we propose a multi-task learning (MTL) framework that jointly predicts the global ADAS-Cog score and its sub-scores (13-items) at Month 24 using baseline MRI and longitudinal clinical scores from baseline and Month 6.
The main goal is to examine how each sub-scores—particularly those associated with MRI features—contribute to the prediction of the global score, an aspect largely neglected in prior MTL studies. We employ Vision Transformer (ViT) and Swin Transformer architectures to extract imaging features, which are fused with longitudinal clinical inputs to model cognitive progression.
Our results show that incorporating sub-score learning improves global score prediction. Subscore-level analysis reveals that a small subset—especially Q1 (Word Recall), Q4 (Delayed Recall), and Q8 (Word Recognition)—consistently dominates the predicted global score. However, some of these influential sub-scores exhibit high prediction errors, pointing to model instability. Further analysis suggests that this is caused by clinical feature dominance, where the model prioritizes easily predictable clinical scores over more complex MRI-derived features.
These findings emphasize the need for improved multimodal fusion and adaptive loss weighting to achieve more balanced learning. Our study demonstrates the value of sub-score-informed modeling and provides insights into building more interpretable and clinically robust AD prediction frameworks. (Github repo provided)
\end{abstract}


\section{Introduction}
According to reports from the Alzheimer’s Disease International (ADI) and the World Health Organization (WHO), the number of people living with dementia is expected to rise significantly by the year 2050, largely due to global population aging \cite{RN23, RN24} and the most common disease highly related is Alzheimer's Disease (AD). Subtle brain structure and neurochemical changes may already exhibit long before clinical symptoms of AD emerge \cite{RN98}. Hence, early detection and prognosis of AD are critical for enabling timely and appropriate interventions that may delay disease progression.

The integration of artificial intelligence (AI) into AD research has catalyzed a surge in studies focused on disease stage classification, leveraging state-of-the-art learning techniques. Classification of disease stages has become a prominent and saturated research area, with most studies focusing on distinguishing structural brain changes among patients with AD, mild cognitive impairment (MCI), and normal controls (NC) \cite{RN84, RN1} 
 based on structural MRI. While these classification approaches have shown satisfactory performance, a binary or trinary classification like these may not offer an accurate representation of the individual’s cognitive state. This limitation presents challenges for developing automated tools for disease progression prognosis.

Prognostic modeling, by contrast, aims to predict clinical cognitive scores, which are more detailed indicators of disease severity. Predicting these scores requires a deeper understanding of AD pathology and its progression over time. Neuropsychological assessments, such as the Alzheimer's Disease Assessment Scale (ADAS-Cog), are known to be strongly correlated with disease status \cite{RN86} as they capture behavioural and cognitive deficits associated with AD. Hence, accurate prediction of clinical scores not only reflects disease progression but also enables better understanding of how cognitive decline is linked to specific brain regions, offering valuable support for disease prognosis.

Graded scores such as ADAS-Cog provide a measure of cognitive function, enabling more detailed assessments and improved sensitivity to subtle cognitive changes, which may be crucial for early AD detection. However, there are relatively fewer studies focused on clinical scores prediction due to regressing continuous variable is a much more challenging task in practice \cite{RN48}, which require extensive predictive model. A critical gap in the literature on AI-based approaches for AD lies in the limited exploration of the relationship between structural brain changes (as observed through structural MRI) and cognitive decline, as quantified by clinical scores such as ADAS-Cog. Most existing studies predominantly emphasizes the prediction of current clinical scores, serving as a benchmark for evaluating feature extraction and disease stage classification. 

In contrast, fewer studies focus on predicting future cognitive scores, which is crucial for early diagnosis and prognosis of the disease \cite{RN16, RN8, RN101, RN102}. Among those that do, the primary focus is often on the prediction of ADAS-Cog global score, while  sub-scores are often overlooked. However, predicting the global score alone tends to be a 'black box' process, raising concerns about whether such models truly learns the structural brain changes related to the disease or establish meaningful links between MRI features and cognitive decline through global scores. This issue is particularly relevant given that ADAS-Cog global scores are conventionally derived as the sum of its sub-scores, each sub-score relates to a different cognitive functions and often to specific areas of the brain. Therefore, neglecting these sub-scores may limit the model’s ability to learn local feature representations and establish robust spatial correspondences between structural MRI and cognitive functions, ultimately degrading both predictive performance and model interpretability.

Recent studies have explored multi-task learning (MTL) approaches, predominantly focusing on prediction of ADAS-Cog global scores or combining classification and regression tasks within a shared learning framework. These models usually adopt MTL formulations such as sparse regression, or hybrid classification–regression networks \cite{RN16, RN8, RN101}.
However, most of these work concentrate only on predicting global scores, which represent an overall aggregation of cognitive function derived from multiple sub-scores, thereby overlooking the spatial and functional differences reflected in each sub-scores. In contrast, sub-scores—such as the 13 items in ADAS-Cog—provide more detailed and fine-grained information about specific cognitive functions (e.g., memory, language, attention).
To date, no study has focused on multi-output regression-based MTL that jointly predicts both the ADAS-Cog global score and its sub-scores (13-items), despite these sub-scores provide critical local feature information that can significantly supports accurate the prediction of the global score.

Addressing this gap, this study aims to to predict future ADAS-Cog 13 global scores at Month 24 using input features comprising baseline MRI as well as ADAS-Cog 13 scores at baseline and Month 6. Our objectives are: (1) develop an MTL framework capable of predicting future global cognitive scores by jointly learning the regression tasks for both global and  ADAS-Cog sub-scores (13-items); (2) investigate the effectiveness of state-of-the-art deep learning architectures—specifically, Vision Transformer (ViT) and Swin Transformer—in extracting meaningful MRI features and accurately forecasting global scores at Month 24; and (3) evaluate the performance of the proposed methodology by comparing it against existing approaches and validated ground truth clinical measurements. This prediction target was selected based on the understanding that individuals may already exhibit subtle structural and neurochemical changes in the brain during this prodromal phase, even before significant clinical symptoms of Alzheimer's disease become apparent.

The rest of the paper is organized as follows. In Section II, we explain our proposed methodology. Section III elaborates on the results and discussion. Finally, Section IV concludes the paper with ideas on possible extensions to this work.

\section{Methodology}

\begin{figure}
    \centering
    \includegraphics[width=1.0\linewidth]{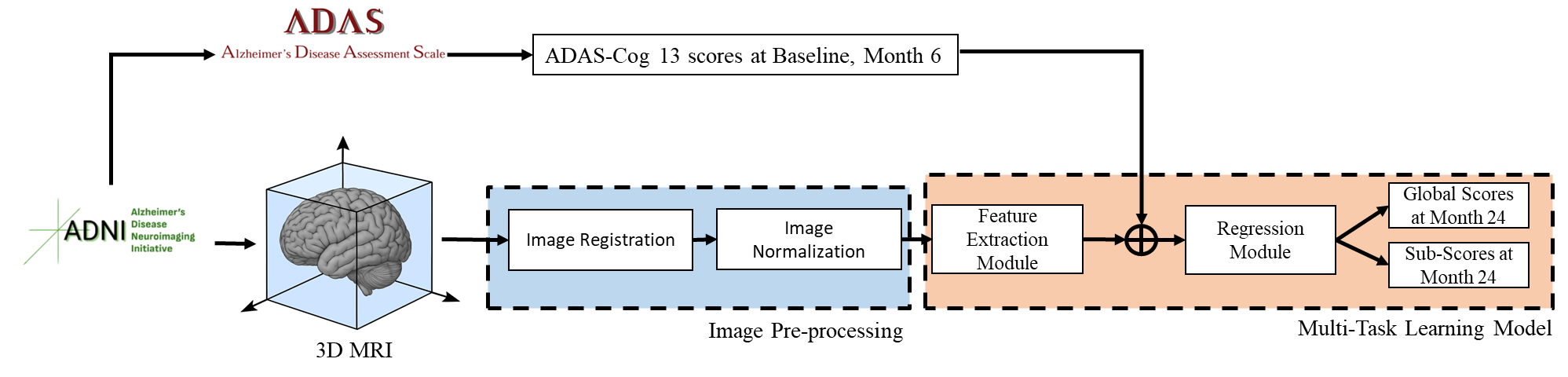}
    \caption{Proposed MTL Framework}
    \label{fig:MTL framework}
\end{figure}

The methodology in this paper involves several key steps as illustrated in Figure \ref{fig:MTL framework}: data acquisition, image preprocessing and multi-task learning model training.

\subsection{Data Acquisition}

In this study, data were obtained from the Alzheimer’s Disease Neuroimaging Initiative (ADNI) database (https://ida.loni.usc.edu/login.jsp), comprising structural MRI (sMRI) scans and neuropsychological test scores from participants diagnosed as cognitively normal (NC), with mild cognitive impairment (MCI), or Alzheimer’s disease (AD).

\begin{table}[h]
    \centering
    \caption{Demographic Characteristics}
    \label{tab:demographics}
    {\large
    \resizebox{1.0\columnwidth}{!}{
    \begin{tabular}{|>{\centering\arraybackslash}p{1cm}|c|c|c|c|}
        \hline
        \multirow{2}{*}{Group} & \multirow{2}{*}{Subject (n)} & \multicolumn{3}{c|}{Demographic Characteristics} \\ \cline{3-5}
        & & Gender (Male/Female) & Age (mean ±SD) & ADAS-Cog (mean±SD) \\ \hline
        AD & 17 & 8/9 & 73.4 ± 8.3 & 17.1 ±2.1 \\ \hline
        NC & 203 & 102/101 & 76.0 ± 5.2 & 9.5 ±4.2 \\ \hline
        MCI & 215 & 138/77 & 74.7 ± 7.5 & 14.5 ±3.9 \\ \hline
    \end{tabular}
    }}

\end{table}

The study focuses on longitudinal data collected at three time points: baseline, 6 months, and 24 months. Subject selection was based on demographic information, cognitive assessments—particularly the Alzheimer's Disease Assessment Scale-Cognitive Subscale (ADAS-Cog)—and MRI acquisition protocols. Specifically, T1-weighted 3D volumetric MPRAGE scans with a sagittal acquisition plane acquired using a 1.5T Siemens scanner were included.

Demographic variables such as age, sex, ADAS-Cog scores, and diagnostic groupings are summarized in Table~\ref{tab:demographics}. This study considers global ADAS-Cog scores ranging from 0 to 20, with a score of 20 indicating moderate Alzheimer's disease, as recommended by \cite{RN71}. Additionally, the dataset includes ADAS-Cog sub-scores (13-items).

\subsection{Image Preprocessing}

A total of 435 baseline MRI scans from ADNI 1 cohort were utilized. or each subject, the ADAS-Cog global scores at Month 24 and all sub-scores (13-items) at baseline, Month 6, and Month 24 were included. Due to anatomical and intensity variations across MRI scans, which can impact consistency and accuracy, two key preprocessing steps were applied to minimize variability and enhance the reliability of subsequent MRI analyses:

\begin{enumerate}[label=\alph*.]

    \item Image registration ensures that the MRI scans are aligned to a common anatomical space,

    \item Image normalization adjusts the image intensities to ensure consistency across scans, which helps reduce inter-subject variability.
\end{enumerate}
Both image registration and normalization were performed using the Advanced Normalization Tools (ANTS) \cite{RN94}.

Image normalization is performed in two-fold:

\begin{enumerate}
    \item \textbf{N4 Bias Field Correction using ANTs:} This step corrects low-frequency intensity non-uniformities arising from scanner-related bias fields. It improves intra- and inter-subject consistency but does not involve any post-hoc intensity evaluation or thresholding to assess normalization success.
    
    \item \textbf{MONAI Intensity Normalization:} After ANTs-based correction, we apply additional intensity normalization using MONAI transforms. This step scales voxel intensity values of each MRI scan independently to a consistent range between 0.0 and 1.0. It further reduces intersubject variability and prepares the data for deep learning input.
\end{enumerate}

\subsection{Proposed Multi-Task Learning Model (MTL)}
Our multi-task learning (MTL) model comprises two main components: (1) a Feature Extraction module and (2) a Regression module. The goal is to perform multiple related regression tasks concurrently—specifically, predicting the future global score and the sub-scores (13-items) of the ADAS-Cog. We adopt the MTL approach based on the understanding that sharing information across tasks enables the extraction of common features from MRI scans and previous clinical scores that are informative for multiple clinical outcomes, thereby enhancing predictive performance and robustness.

\subsubsection{Feature Extraction Module}

Feature extraction is a critical step in which salient information or representative attributes are derived from MRI data. In this work, we investigated two transformer-based architectures—Vision Transformer (ViT) and Swin Transformer —due to their capacity to efficiently model global and local relationships across an image without relying on a fixed receptive field \cite{RN75}. This capability is particularly advantageous in the medical imaging domain, where capturing global context is essential for accurate disease diagnosis.

\subsubsection{Regression Module}

Our regression module is designed to jointly learn two related group of tasks: (1) regression of the ADAS-Cog global scores at Month 24 (2) regression of ADAS-Cog sub-scores (13-items) at Month 24. While conceptually grouped into global and sub-score predictions, the overall setup involves 14 distinct regression tasks. Following the feature extraction step, the shared information is passed through task-specific fully connected layers, i.e., one for each of the 14 outputs. This structure allows all tasks to benefit from shared features learned in the backbone, while still allowing task-specific fine-tuning through their respective output layers. We define a shared loss function that combines the mean squared errors (MSE) from both regression tasks to facilitate joint learning. A combined loss function is defined by aggregating the mean squared errors (MSE) from all 14 regression tasks, as shown in Equations 1–3, to support effective joint learning.

\begin{equation}
\mathcal{L}_{\text{total}}(X, Y) = \alpha \cdot \mathcal{L}_{\text{mse (sub-scores)}} + (1 - \alpha) \cdot \mathcal{L}_{\text{mse (global)}}
\label{eq:total_loss}
\end{equation}
\noindent
where $\alpha \in [0, 1]$ is a weighting factor that balances the contributions of the sub-scores and global score regression tasks.

\begin{equation}
\mathcal{L}_{\text{mse (global)}} = \frac{1}{n} \sum_{i=1}^{n} \left( y_i - \hat{y}_i \right)^2
\label{eq:global_loss}
\end{equation}
\noindent
where $y_i$ and $\hat{y}_i$ denote the ground truth and predicted ADAS-Cog global scores for the $i^{th}$ sample, and $n$ is the number of samples.

\begin{equation}
\mathcal{L}_{\text{mse (sub-scores)}} = \frac{1}{13} \sum_{j=1}^{13} \left( \frac{1}{n} \sum_{i=1}^{n} \left( y_{ij} - \hat{y}_{ij} \right)^2 \right)
\label{eq:subscore_loss}
\end{equation}
\noindent
where $y_{ij}$ and $\hat{y}_{ij}$ represent the ground truth and predicted values of the $j^{th}$ ADAS-Cog sub-score for the $i^{th}$ sample.

\subsection{Model Training and Evaluation}

We trained our multi-task learning (MTL) model using preprocessed and normalized baseline MRI scans to predict future ADAS-Cog global scores and sub-scores (13-items) at Month 24. Additionally, we investigated the impact of aggregating features from baseline MRI, and ADAS-Cog sub-scores (13-items) at baseline and Month 6 on the model's performance in predicting future clinical scores.
The training process involved the following steps:
\begin{enumerate}
    \item Data Splitting: The dataset was divided into training and validation sets at the subject level to prevent data leakage and ensure robust model evaluation.
    \item Model Training: The MTL model was trained on the training set using an Adam optimizer and a combined loss function tailored for multi-output regression. We set the weighting factor $\alpha = 0.5$to assign equal importance to the tasks—predicting the global ADAS-Cog score and the subscores (13-items)—so that the model learns them in a balanced manner. The idea is to prevent the model from being biased toward either task during training.
    \item Evaluation Metrics: Model performance was assessed using mean absolute error (MAE), root mean squared error (RMSE), and Pearson correlation coefficient (r). The Pearson correlation 
    as used to quantify the linear relationship between the actual and predicted scores, with higher values indicating stronger predictive performance.
	\item Explanability using SHapley Additive exPlanations (SHAP): Using Explainable AI model, SHAP \cite{RN103}, we figured out the features that importance to identify the most influential features driving the prediction of ADAS-Cog global scores and sub-scores.
\end{enumerate}

\section{Results and Discussion}

\subsection{Global Score at Month 24 Prediction}

\begin{table}[h]
    \centering
    \caption{MTL Framework: Performance evaluation based on utilization of (1) combination anatomical + cognitive input data; (2) anatomical input data.}
    \label{tab:placeholder_label}
    {\large
    \resizebox{1.0\columnwidth}{!}{
    \begin{tabular}{|>{\centering\arraybackslash}p{3cm}|>{\centering\arraybackslash}p{2cm}|>{\centering\arraybackslash}p{4cm}|r|r|>{\centering\arraybackslash}p{1cm}|}
    \hline
    \textbf{Feature Extraction Module} & \textbf{Regression Module} & \textbf{Input data} & \textbf{MAE} & \textbf{RMSE} & \textbf{r} \\ \hline
    N/A & Linear Layer & ADAS-Cog clinical scores & 2.51 & 3.31 & 0.78 \\ \hline
    Swin Transformer (Ours) & Linear Layer & Baseline MRI + ADAS-Cog clinical scores & 2.60 & 3.45 & 0.76 \\ \cline{3-6} 
    & & Baseline MRI & 4.65 & 5.32 & -0.06\\ \hline
    ViT (Ours) & Linear Layer & Baseline MRI + ADAS-Cog clinical scores & 2.53 & 3.30 & 0.79 \\ \cline{3-6} 
    & & Baseline MRI & 4.64 & 5.31 & 0.08 \\ \hline
    Dirty Model \cite{RN101} & Linear Layer & Baseline MRI & 3.18 & N/A & 0.08 \\ \hline
    \end{tabular}
    }}
\end{table}

Table \ref{tab:placeholder_label} summarizes the performance metrics of the MTL model using different feature extraction backbones with a consistent regression module composed of linear layers. An ablation study was performed with three input configurations: (1) ADAS-Cog clinical scores only, (2) baseline MRI only, and (3) both modalities combined. The combined modalities setting yielded the best performance—MAE: 2.53 (ViT), 2.60 (Swin); Pearson: 0.79 and 0.76, respectively—demonstrating the benefit of integrating anatomical and cognitive inputs.

To ensure a fair comparison with Imani et al. \cite{RN101}, we reconfigured our model to use baseline MRI-only input. The resulting ViT-based model showed reduced performance, particularly a low Pearson correlation, though it remained comparable to the "Dirty Model" of \cite{RN101}. This suggests limited learning from anatomical MRI data alone. Conversely, the clinical scores-only model achieved superior performance (MAE: 2.51; Pearson: 0.78), indicating that (1) clinical score features alone provide highly predictive information for future globasl scores, and (2) the model exhibits a learning bias toward clinical scores features. Furthermore, clinical scores-only models performed similarly to the MTL model that uses both MRI and clinical data, whereas MRI-only models showed significant performance degradation. These findings highlight a strong modality imbalance, suggesting that the model underutilizes MRI-derived information.

As we applied modality-dropout (i.e., using MRI-only features without clinical data), the model exhibited underperformance, notably a drastic drop in Pearson correlation. Several factors contribute to this:

\begin{enumerate}[label=\alph*.]
    \item \textbf{Low Variability in Structural MRI Among NC Group} \\
    The NC group inherently shows minimal anatomical changes, leading to low variability in structural MRI data. This makes it difficult for the model to learn meaningful patterns, resulting in poor performance—particularly in distinguishing subtle score variations.
    
    \item \textbf{Imbalanced and Small Sample Size} \\
    The dataset suffers from class imbalance and limited sample size. This further restricts the model’s ability to generalize and predict accurately, especially when deprived of auxiliary clinical features.
    
    \item \textbf{Ineffectiveness of Uniform Balanced Weighting ($\alpha$)} \\
    Although we applied balanced weighting ($\alpha$) across all ADAS-Cog sub-scores during training, this non-adaptive uniform approach seems unsuitable for MRI-only input. Certain sub-scores—such as Q1 (Word Recall), Q4 (Delayed Word Recall), and Q8 (Word Recognition)—are more directly linked to brain structural changes observable in MRI. These sub-tasks likely require stronger supervision, and their losses should be penalized more heavily to guide the model effectively.
\end{enumerate}

Hence, adaptively weighting the loss based on task-specific MRI sensitivity will be the focus of our future investigation.

\begin{table}[h]
\centering
\caption{SHAP Value Comparison: Feature Importance based on MTL}
\label{tab:shap_results}
\begin{tabular}{|c|c|}
\hline
\textbf{ViT} & \textbf{Swin Transformer} \\
\hline
\includegraphics[width=0.22\textwidth]{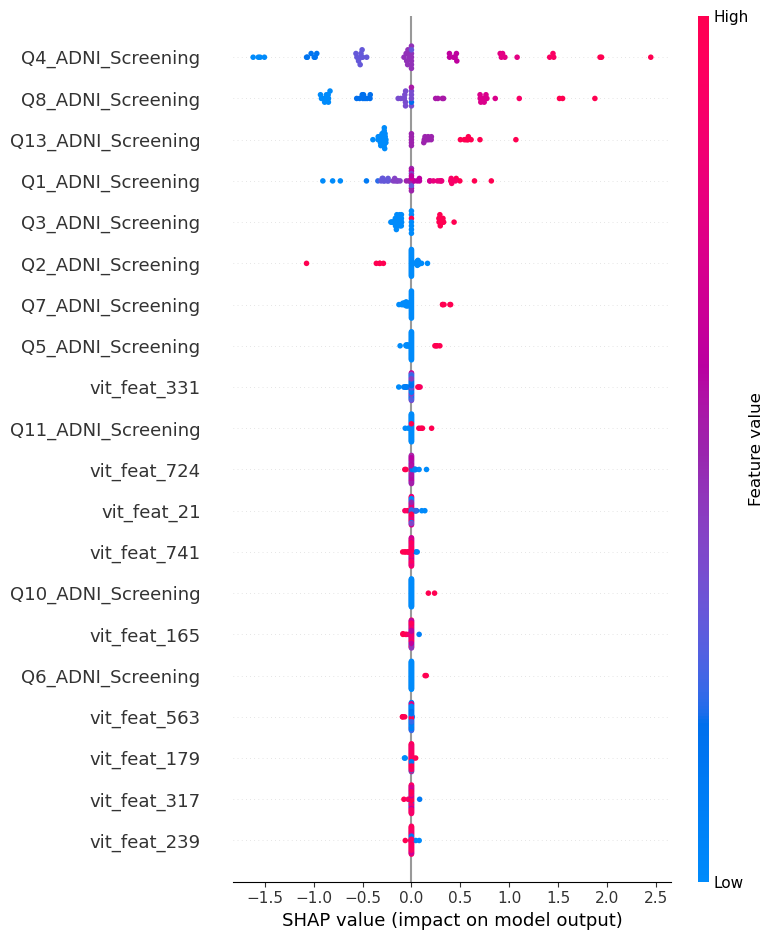} & \includegraphics[width=0.22\textwidth]{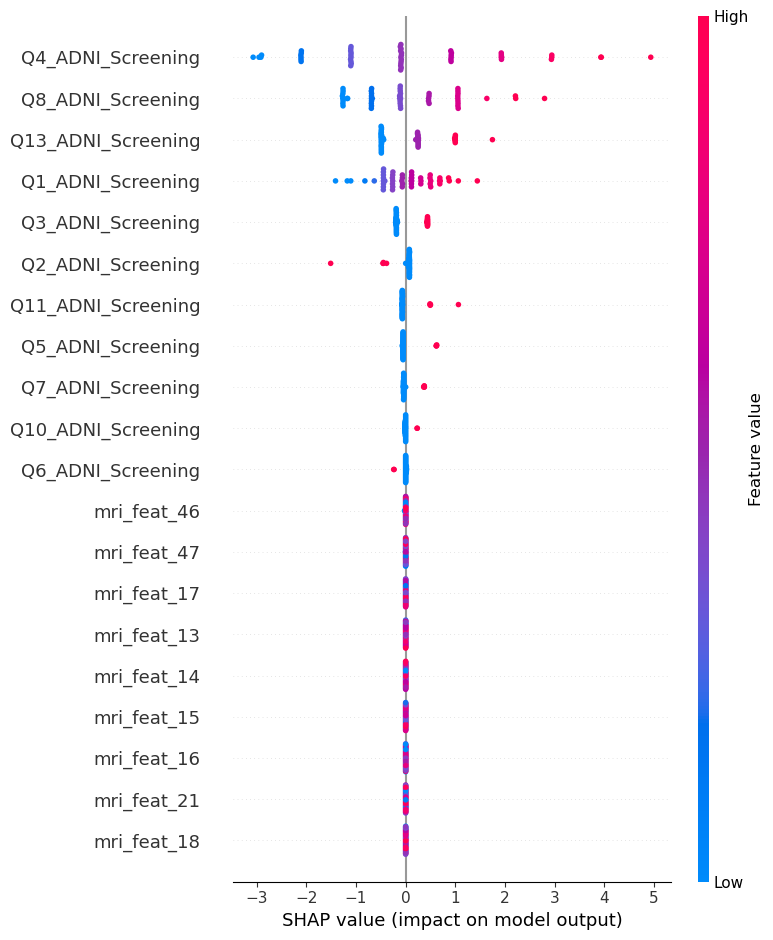} \\
\hline
\end{tabular}
\end{table}

SHAP analysis as illustrated in Table \ref{tab:shap_results} revealed that clinical scores features, particularly ADAS-Cog sub-scores at baseline and Month 6—consistently ranked highest in importance across both tasks (global and sub-score prediction), while ViT-derived MRI features showed negligible contribution. 

While the use of both MRI and clinical data was intended to enhance prediction, our findings suggest that the model predominantly learns from clinical scores inputs. This is likely due to the strong direct correlation between past and future ADAS-Cog scores, making clinical scores data an easier signal to optimize. As a result, the model may underexploit the rich spatial information embedded in MRI. This raises important concerns about interpretability and generalizability, particularly in scenarios where clinical scores may be unavailable or unreliable.

Given that multi-task learning (MTL) facilitates shared representation across related regression tasks, we posit that the prediction of the global ADAS-Cog score is inherently influenced by its underlying 13 sub-scores. By jointly predicting both the global score and the sub-scores (13-items), our model leverages local feature representations and establishes robust spatial correspondences between structural MRI and specific cognitive functions reflected in each sub-score. 

\begin{table}[ht]
    \centering
    \caption{Subscore-Level Evaluation for MCI and NC Subjects}
    \label{tab:subscore_evaluation}
    {\large
    \resizebox{1.0\columnwidth}{!}{
    \begin{tabular}{|l|>{\centering\arraybackslash}p{2.5cm}|>{\centering\arraybackslash}p{2cm}|>{\centering\arraybackslash}p{2.5cm}|>{\centering\arraybackslash}p{2.5cm}|>{\centering\arraybackslash}p{1.5cm}|c|c|>{\centering\arraybackslash}p{2.5cm}|}
    \hline
    \textbf{Subject} & \textbf{Predicted Global Score} & \textbf{Subscore} & \textbf{True Subscore} & \textbf{Predicted Subscore} & \textbf{Abs Error} & \textbf{MAE} & \textbf{RMSE} & \textbf{Contribution (\%)} \\
    \hline
    MCI & 8.37 & Q1 & 0.67 & 1.11 & 0.44 & 0.44 & 0.663 & 31.15 \\
        &      & Q4 & 0.67 & 0.04 & 0.63 & 0.63 & 0.793 & 1.19 \\
        &      & Q8 & 0.00 & 1.44 & 1.44 & 1.44 & 1.200 & 40.36 \\
    \hline
    NC  & 14.42 & Q1 & 0.00 & 0.13 & 0.13 & 0.13 & 0.361 & 30.36 \\
        &       & Q4 & 0.00 & 0.47 & 0.47 & 0.47 & 0.685 & 30.23 \\
        &       & Q8 & 0.00 & 0.04 & 0.04 & 0.04 & 0.200 & 24.30 \\
    \hline
    \end{tabular}
    }}
\end{table}

To evaluate this hypothesis, we conducted a two-part analysis on subjects from two clinical categories (i.e., MCI and NC) (See Table \ref{tab:subscore_evaluation}):
\begin{enumerate}
    \item \textbf{Influence on Global Prediction:} We quantified how much each subscore influenced the predicted global score by computing the percentage contribution of each predicted subscore (Q1–Q13) relative to the total predicted global score. Across both subjects, three sub-scores—Q1 (Word Recall), Q4 (Delayed Word Recall), and Q8 (Word Recognition)—consistently accounted for over 80(\%) of the predicted global score. This suggests that the model internally prioritizes specific cognitive domains when forming its global prediction.

    \item \textbf{Subscore Predictive Accuracy:} We further evaluated whether the model’s reliance on these high-impact sub-scores was justified by examining their prediction performance using MAE, RMSE and absolute error. For the NC subject, these sub-scores not only contributed the most to the predicted global score but also exhibited low error values, indicating accurate and stable learning. In contrast, for the MCI subject, Q4 and Q8 contributed significantly to the global score but showed substantial prediction errors, particularly for Q8 (abs error = 1.44), likely due to model instability or insufficient representation learning.
\end{enumerate}

These findings reinforce the value of our MTL formulation: the global score prediction is not derived in isolation but emerges through a learned aggregation of cognitively meaningful sub-scores. They also highlight the importance of examining sub-score predictions when interpreting model outputs (i.e. global scores). Notably, the global score is not uniformly shaped by all 13 sub-scores, but rather by a few key sub-scores (components). This insight opens up opportunities to adjust the loss weighting (alpha) in future work — giving more focus to subscores that are both impactful and MRI-relevant, to encourage more balanced and interpretable learning.


\section{Conclusion}

This study presented a multi-task learning (MTL) framework for predicting future ADAS-Cog 13 global scores at Month 24 by jointly modeling both the global score and its sub-scores (13-items). Unlike most previous studies that focus solely on predicting the global ADAS-Cog score, our approach introduces simultaneous subscore prediction to capture both coarse and fine-grained cognitive changes. This design enables richer clinical insights and improves interpretability by revealing how specific cognitive domains influence the overall trajectory of decline.

By employing transformer-based architectures (ViT and Swin Transformer), we combined structural MRI features with clinical scores from baseline and Month 6 to model disease progression. Our findings demonstrate that incorporating sub-scores learning improves global score prediction and provides deeper insight into the multidimensional nature of cognitive impairment.

Through the analysis, we observed that a small set of sub-scores, particularly Q1 (Word Recall), Q4 (Delayed Word Recall), and Q8 (Word Recognition)—consistently contributed the most to the predicted global score. However, some of these sub-scores also exhibited high prediction errors, indicating model instability. This instability appears to stem from clinical feature dominance, caused by (1) the strong direct correlation between past and future clinical scores and (2) underutilization of MRI features due to suboptimal feature integration.

These results suggest that the global score is not evenly influenced by all sub-scores, and that rebalancing the training focus may improve learning from MRI-relevant sub-scores. Future work should explore improved multimodal fusion strategies and adaptive weighting to reduce over-reliance on clinical inputs and enhance the robustness and generalizability of MTL-based AD prediction models and promote more balanced multimodal learning through architecture-level modifications.

\bibliographystyle{ieeetr}
\bibliography{references}

\end{document}